\documentclass[letterpaper, 10 pt, conference]{ieeeconf}  

\IEEEoverridecommandlockouts                              

\usepackage{graphics} 
\usepackage{epsfig} 
\usepackage{mathptmx} 
\usepackage{times} 
\usepackage{amsmath} 
\usepackage{amssymb}  
\usepackage{multicol}
\usepackage{color,soul}

\title{\LARGE \bf Haptic Sketches on the Arm for Manipulation in Virtual Reality}

\author{Mine Sarac$^{1}$, Allison M. Okamura$^{1}$, and Massimiliano Di Luca$^{2}$
\thanks{$^{1}$ Department of Mechanical Engineering, Stanford University,  Stanford, CA.
        {\tt\small msarac@stanford.edu, aokamura@stanford.edu}}%
\thanks{$^{2}$ Facebook Reality Labs, and University of Birmingham, UK
        {\tt\small max.diluca@oculus.com}}%
}

\begin{document}

\maketitle
\thispagestyle{empty}
\pagestyle{empty}

\begin{abstract}

We propose a haptic system that applies forces or skin deformation to the user's arm, rather than at the fingertips, for believable interaction with virtual objects as an alternative to complex thimble devices. Such a haptic system would be able to convey information to the arm instead of the fingertips, even though the user manipulates virtual objects using their hands. We developed a set of haptic sketches to determine which directions of skin deformation are deemed more believable during a grasp and lift task. Subjective reports indicate that normal forces were the most believable feedback to represent this interaction. 

\end{abstract}
\section{Introduction}

\begin{figure}[t!]
  \centering
  \resizebox{2.8in}{!}{\includegraphics{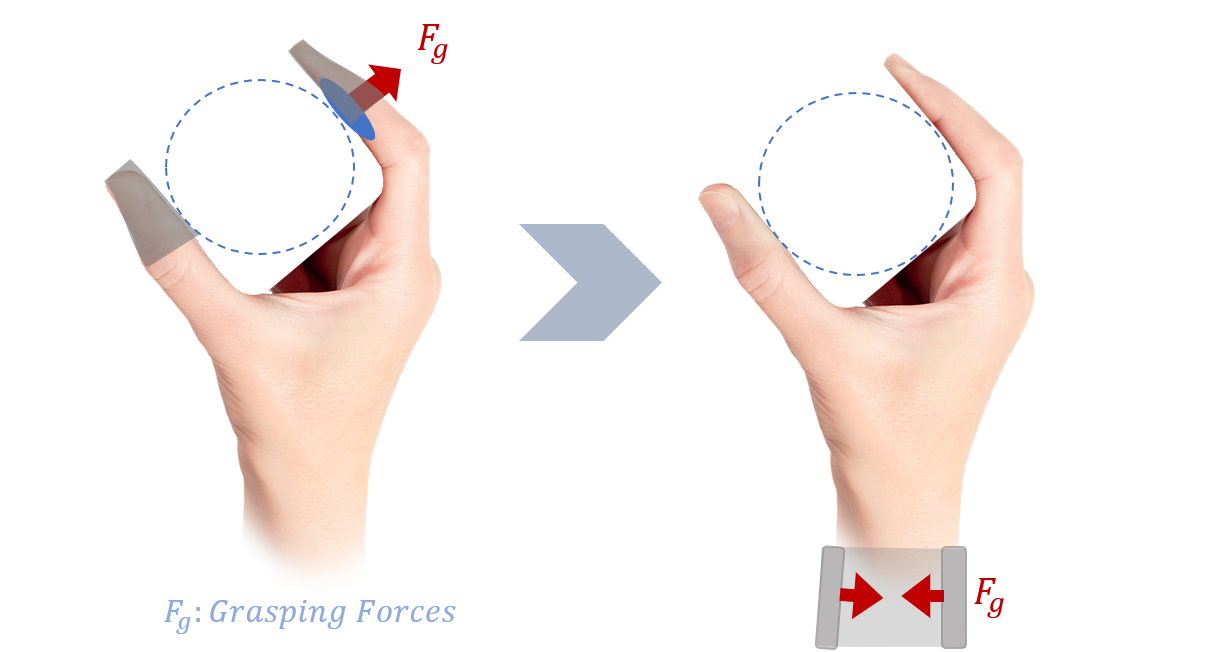}}
  \caption{Haptic feedback modalities: (a) current haptic devices where a fingertip device conveys grasping ($F_g$) forces, and (b) the proposed haptic system where grasping ($F_g$) forces are rendered at the wrist or arm.}
  \label{fig:proposed}
  \vspace*{-1\baselineskip}
\end{figure}

In the real world, mechanical properties of objects, such as mass, stiffness and temperature, are mostly sensed through touch. 
Haptic devices aim to recreate the same feeling for virtual interactions, typically at the fingertip. 
Multi-degree-of-freedom fingertip devices (Fig.~\ref{fig:proposed}~(a)) provide skin deformation for compelling and realistic feedback~\cite{Suchoski2018, Leonardis2017}. 
Even though these multi-degree-of-freedom devices improve performance and perceived realism of manipulation tasks, they are complex, bulky, and expensive.

We propose a different approach to artificial haptic feedback, by relocating the haptic feedback that conveys information about mechanical properties of manipulated virtual objects from the fingertip to the arm (Fig.~\ref{fig:proposed}~(b)). Rather than recreating realistic feedback to match the stimulation of real objects, we posit that sensory signals, which merely hint at the real properties of an object, might be sufficient to create interpretable or ``believable" interactions. Such a concept has been applied, for example, in pseudo-haptic rendering, where the visual analogies are used to inform the user of what haptic effects they would have experienced. 

The typical requirement of realism turns into the new requirement of believability, if haptic feedback accompanies virtual manipulation tasks with visual cues. In this context, a believable haptic feedback can convey information about material properties of objects to be useful for manipulation tasks, so that it qualitatively adds to (rather than detracts from) the user experience.

Prior work has examined methods for haptic feedback at the arm. 
For example, ``CUFF'' applies normal and tangential forces to render interaction forces between objects in the environment, and a hand prosthesis, which is worn by the same user who wears the ``CUFF''~\cite{Casini2015}. 
``hBracelet'' also applies normal and tangential forces to render interaction forces that occur during virtual manipulation or teleoperation tasks~\cite{Meli2018}. 
Moriyama et al.~\cite{Moriyama2018} also developed a haptic device with two five-bar linkage mechanisms, applying normal and tangential forces, to represent interaction forces acting on the index finger and the thumb independently during virtual grasping tasks.
Even though these devices were developed with different motivations, they share the mutual goal (and associated design complexity) of creating realistic feedback for the user.

Simpler haptic devices could instead be designed to render believable information to the user about the physical properties of virtual objects during manipulation tasks. 
To investigate the use of skin stretch applied at the wrist as a way to create believable sensations, we investigated which directions of skin deformation on a user's arm would result in the most believable feedback.
We created a set of haptic sketches in which feedback forces (applied via skin deformation) were stimulated manually. 
Here we present these haptic sketches and our findings, and discuss future work toward designing a wearable haptic device.

\section{Haptic Sketches for Virtual Manipulation}


We investigated how users perceived interactions during virtual manipulation tasks presented via skin deformation on the arm, while the users viewed the manipulation task in the virtual environment. We implemented simple haptic sketches, which were introduced by Moussette~\cite{Moussette2012}. Haptic sketches aim to stimulate feedback forces manually or through simple tools (like magnets) instead of complex mechanical devices. 
Even though applying forces manually might not correlate with the virtual interactions accurately, the repetition of these tasks might result in sufficient believability to compare among different feedback modalities. Furthermore, removing the need for a mechatronic device gave us the opportunity to quickly test several types of stimulation.



\begin{figure}[t!]
  \centering
  \resizebox{2.9in}{!}{\includegraphics{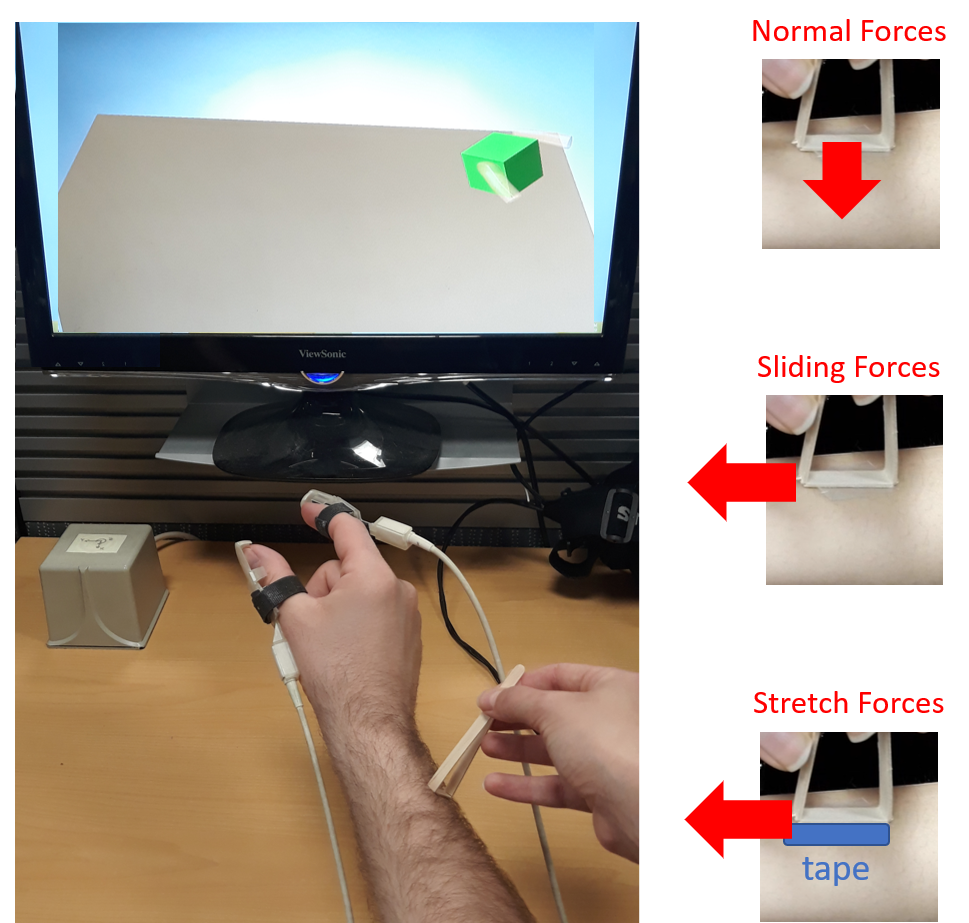}}
  \caption{Haptic sketches: (a) the setup with a virtual environment, two motion trackers to be worn at the fingertips on the index and middle fingers, and a folded wooden stick to stimulate haptic feedback manually by the experimenter, and (b) different haptic feedback modalities applied to the user, such as normal, sliding and stretch forces. Skin stretch forces are implemented using a double-sided tape between the stirrer and user's skin.}
  \label{fig:sketch_env}
  \vspace*{-1\baselineskip}
\end{figure}

Fig.~\ref{fig:sketch_env} shows the setup we employed for haptic sketches. During trials, a user was asked to sit in front of a monitor and wear electromagnetic finger trackers on the thumb and index finger. On the monitor, the user sees a representation of a virtual box and their virtual fingertips, whose movements are mapped to the actual fingers. The user is asked to grasp the object and move it around. In the meantime, the experimenter applies feedback in the form of skin stretch to the user's arm at a single point using a folded wooden stick (made from a coffee stir stick). The stick is folded into a U-shape so that the middle area is used as the contact area, and the open extremities are used as a handle by the experimenter to apply forces. For normal and sliding forces, the stick contacts the user's skin directly, while for stretch forces, double-sided tape is used. The experimenter aims to provide feedback forces to reflect the stiffness of the object being grasped.


The haptic sketches were tested on five subjects, all students who are familiar with haptics. Each user experienced all feedback modalities repeatedly until they felt confident with the perception. 
After each feedback modality, subjects were asked whether they could feel the feedback throughout the whole manipulation task, how they felt about its correspondence with the manipulation task, and whether they could learn to interpret the mechanical properties of objects with it. Finally, they were asked to compare all modalities. 


All five users reported the same experience. Such an agreement convinced us that we have a reliable approach for future research. The results suggest that normal forces and stretch forces led to a believable experience of material properties of objects, while sliding sensations were not perceived to be related to manipulation actions or mechanical properties in the virtual environment. Even though users reported that they could perceive mechanical properties of the objects using both normal and stretch forces, normal forces felt more realistic. It is possible that users relate normal forces to natural muscle contraction that occurs as they move their fingers. 

Even though normal and stretch forces might have similar impact on user perception, they are different in terms of mechanical implementation. Skin stretch highly depends on how the end-effector of the haptic device is attached to the user's skin; we need to prevent it from slipping. Even though two-sided tape or glue can be used to attach mechanical parts to the skin, its practicality as well as implementation on hairy skin might be problematic. Furthermore, the mechanical parts need to be grounded on the arm, which results in reaction or grounding forces that also affect user perception. Yet, normal forces are expected to be much simpler to implement. 

As a result of these haptic sketches and implementation considerations, we conclude that a believable haptic device should apply normal forces as users manipulate virtual objects and view their actions in a virtual environment. 

\section{Future Work}

Having decided to apply normal forces during virtual manipulation tasks, we will implement a mechanical device for further investigations. Even though our haptic sketches applied forces to a single area on the surface of the user's arm, we expect that perception can be improved if normal forces are applied at multiple contact points or different locations. The performed haptic sketches focused on stiffness properties of the object, since the forces were applied based on the finger movement with which the user grasped the object. In future implementations, the haptic device could also render inertial forces based on how heavy the object is and the manner in which users lift it. 

In summary, we propose that normal force on the arm could provide information that is sufficient to make a manipulation task occurring in the virtual environment believable. However, even the best implementation of this idea might be insufficient, due to the lack of feedback at the fingertip. In this scenario, we could introduce a simple fingertip device, which can render event-based feedback to represent contact events only and compare the performance of the overall system to existing multi-degree-of-freedom fingertip devices. 


%

\begin{thebibliography}{99}


\bibitem{Suchoski2018} J. M. Suchoski, S. Martinez and A. M. Okamura, ``Scaling Inertial Forces to Alter Weight Perception in Virtual Reality," IEEE International Conference on Robotics and Automation, 2018, pp.484-489.

\bibitem{Leonardis2017} D. Leonardis, M. Solazzi, I. Bortone and A. Frisoli, ``A 3-RSR Haptic Wearable Device for Rendering Fingertip Contact Forces," in IEEE Transactions on Haptics, vol. 10, no. 3, pp. 305-316, 2017.

%
%



%
\bibitem{Casini2015} S. Casini, M. Morvidoni, M. Bianchi, M. Catalano, G. Grioli and A. Bicchi, ``Design and realization of the CUFF - clenching upper-limb force feedback wearable device for distributed mechano-tactile stimulation of normal and tangential skin forces," IEEE/RSJ International Conference on Intelligent Robots and Systems, Hamburg, 2015, pp. 1186-1193.

\bibitem{Meli2018} L. Meli, I. Hussain, M. Aurilio, M. Malvezzi, M. K. O’Malley and D. Prattichizzo, ``The hBracelet: A Wearable Haptic Device for the Distributed Mechanotactile Stimulation of the Upper Limb," in IEEE Robotics and Automation Letters, vol.3, no.3, pp. 2198-2205, 2018.

\bibitem{Moriyama2018} T. K. Moriyama, A. Nishi, R. Sakuragi, T. Nakamura and H. Kajimoto, "Development of a wearable haptic device that presents haptics sensation of the finger pad to the forearm," 2018 IEEE Haptics Symposium (HAPTICS), San Francisco, CA, 2018, pp. 180-185.





\bibitem{Moussette2012} C. Moussette. ``Simple haptics: Sketching perspectives for the design of haptic interactions." PhD dissertation, Umea University, 2012.



\end{thebibliography}


\end{document}